\documentclass[letterpaper, 10 pt, conference]{ieeeconf}  

\IEEEoverridecommandlockouts    

\usepackage{graphics} 
\usepackage{epsfig} 
\usepackage{times} 
\usepackage{amsmath} 
\usepackage{amssymb}  
\usepackage{hyperref}
\usepackage{multirow} 
\usepackage{cite}
\usepackage{xcolor}

\usepackage{booktabs}
\usepackage[ruled,vlined,linesnumbered]{algorithm2e}

\title{\LARGE \bf
GUIDE: A Diffusion-Based Autonomous Robot Exploration Framework Using Global Graph Inference
}

\author{Zijun Che$^{1*}$, Yinghong Zhang$^{1*}$, Shengyi Liang$^{1}$, Boyu Zhou$^{2}$, Jun Ma$^{1}$, Jinni Zhou$^{1}$
\thanks{This work was supported by the Guangdong Provincial Educational Science Planning Project (Grant No. 2025GXJK0629), Guangzhou Higher Education Teaching Research and Reform Project (Grant No. 2024YBJG094), and Guangzhou Municipal Educational Science Planning Project (Grant No. 202316577). \textit{(Zijun Che and Yinghong Zhang contributed equally to this work.) (Corresponding Author: Jinni Zhou.)}}
\thanks{$^{1}$The Hong Kong University of Science and Technology (Guangzhou), Guangzhou, China.
        {\tt\small \{zche624, yzhang780, sliang318\}@connect.hkust-gz.edu.cn, jun.ma@ust.hk, eejinni@hkust-gz.edu.cn }}%
\thanks{$^{2}$Southern University of Science and Technology, Shenzhen, China.
        {\tt\small zhouby@sustech.edu.cn}}%
}

\begin{document}

\maketitle
\thispagestyle{empty}
\pagestyle{empty}


\begin{abstract}

Autonomous exploration in structured and complex indoor environments remains a challenging task, as existing methods often struggle to appropriately model unobserved space and plan globally efficient paths.
To address these limitations, we propose GUIDE, a novel exploration framework that synergistically combines global graph inference with diffusion-based decision-making.
We introduce a region-evaluation global graph representation that integrates both observed environmental data and predictions of unexplored areas, enhanced by a region-level evaluation mechanism to prioritize reliable structural inferences while discounting uncertain predictions.
Building upon this enriched representation, a diffusion policy network generates stable, foresighted action sequences with significantly reduced denoising steps. Extensive simulations and real-world deployments demonstrate that GUIDE consistently outperforms state-of-the-art methods, achieving up to 18.3\% faster coverage completion and a 34.9\% reduction in redundant movements.

\end{abstract}
\section{INTRODUCTION}

Autonomous exploration remains a cornerstone of modern robotics research, with pivotal applications in scenarios where unknown environment coverage is critical: environmental monitoring, warehouse logistics, and search-and-rescue operations~\cite{cave-large,de2022rmf}. A defining challenge in these tasks is efficiently covering all reachable areas under stringent constraints—limited time, finite energy, and constrained computational resources. Despite decades of progress, existing exploration strategies still struggle to appropriately model unobserved space and plan globally efficient paths.

\begin{figure}[htbp]
\vspace{5pt}
\centering
\includegraphics[width=0.48\textwidth]{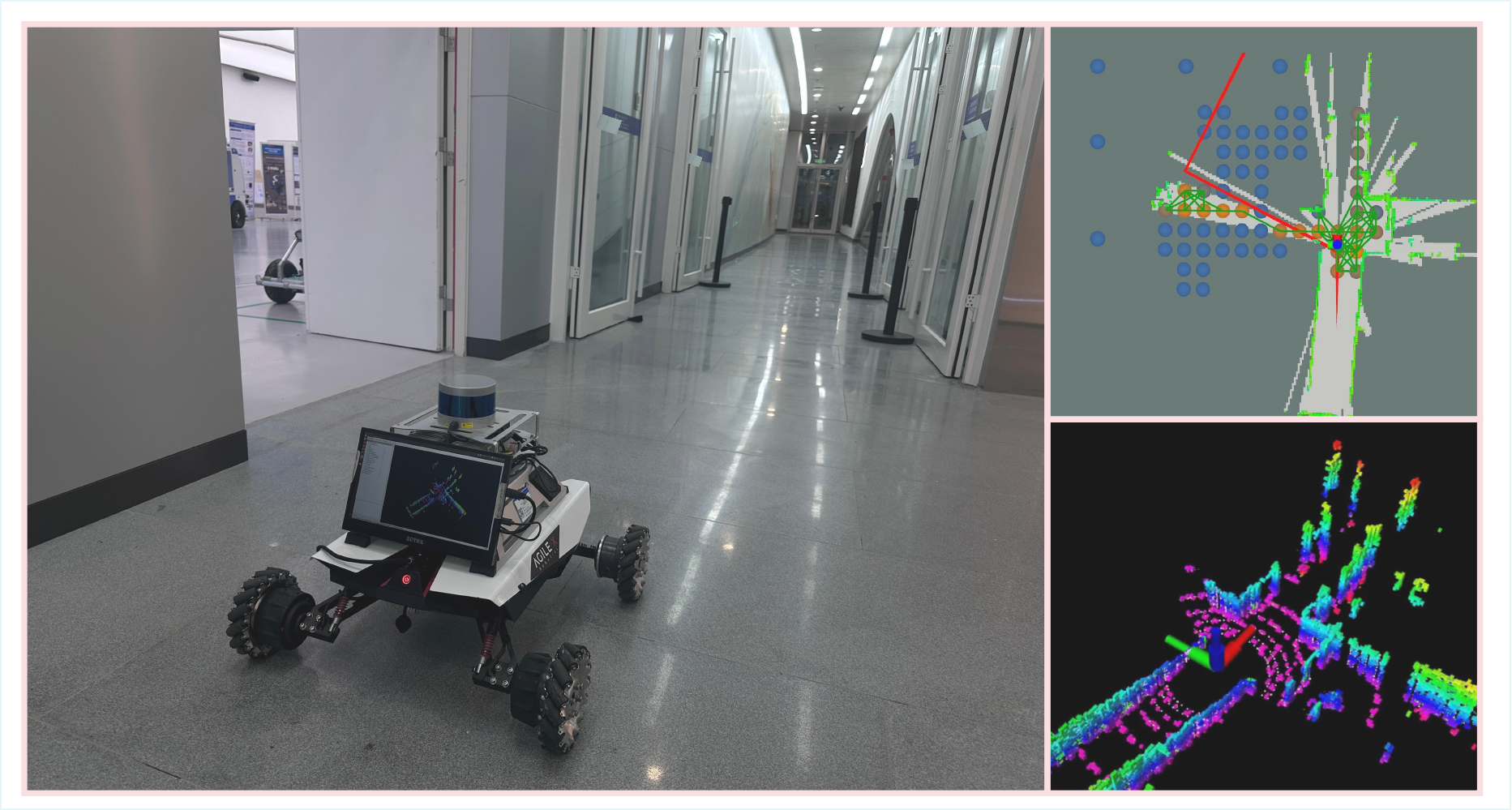} 
\caption{Exploration in a structured indoor environment with a mobile robot using GUIDE. Left: the robot’s actual position in the real world. Top right: the robot leverages the global graph representation to generate exploration trajectories, where blue nodes indicate predictions of unexplored areas and the red line denotes the generated action sequence. Bottom right: the 3D occupancy map constructed by the robot during exploration.}
\label{fig:real}
\end{figure}

Current exploration methodologies can be broadly categorized into model-based and learning-based approaches, each exhibiting fundamental limitations in addressing the global coverage challenge. Early model-based techniques, including frontier-based methods~\cite{zhou2021fuel,cao2021tare} and sampling-based exploration strategies~\cite{nbvp}, rely exclusively on observed map information to determine exploration directions. While these approaches demonstrate reasonable performance in structured environments, their inherent myopia toward unobserved areas frequently results in redundant revisits, inefficient path planning, and suboptimal coverage—particularly in environments with complex topologies. Coverage path-based methods~\cite{kan2020online,zhang2024falcon} attempt to address this limitation by incorporating explicit coverage objectives; however, their reliance on uniform grid decomposition implicitly assumes environmental regularity, leading to performance degradation in spaces with irregular layouts or varying structural complexity.

More recently, learning-based approaches have emerged as promising alternatives, yet they too face significant challenges in achieving comprehensive spatial understanding. The first category employs neural networks to directly map observed environments to exploration actions~\cite{cao2023ariadne,cao2024deep,cao2024dare}. Although these methods improve adaptability to specific environments, they fundamentally operate with limited information—encoding only observed areas while remaining agnostic to the structure of unknown spaces. This inherent constraint severely limits their capacity to achieve globally efficient exploration, often requiring extensive training across diverse environments to achieve moderate performance. The second category explicitly predicts unobserved areas and associated information gain~\cite{tao2022seer,ho2024mapex}. While conceptually promising, these approaches typically utilize predicted maps only for local planning rather than incorporating them into a comprehensive global planning framework, thereby failing to fully leverage the predictive information for long-horizon path optimization.

These limitations collectively highlight a critical research gap: the absence of a unified framework that effectively integrates predictions of unknown areas with globally optimized exploration planning. Specifically, existing methods lack mechanisms to (1) construct a comprehensive environmental representation that coherently combines observed information with predictions of unexplored areas, (2) leverage credible predictions to guide exploration decisions, and (3) generate stable, long-horizon trajectories that maximize coverage efficiency while minimizing redundant movements.

To address these challenges, we propose \textbf{GUIDE}, a novel exploration framework that synergistically combines global graph inference with diffusion-based decision-making. At its core, GUIDE constructs a region-evaluation global graph representation that integrates both observed environmental data and predictions of unexplored areas. This representation is enhanced through a region-level evaluation mechanism that prioritizes significant regional structural inferences, effectively creating an informative yet compact environmental model that prioritizes credible structural inferences while appropriately discounting uncertain predictions. Building upon this enriched representation, GUIDE employs a diffusion policy network that generates stable, foresighted action sequences with significantly reduced denoising steps compared to conventional approaches—enabling efficient long-horizon planning that effectively balances immediate information gain with comprehensive coverage objectives. The reduced computational overhead ensures real-time responsiveness—a critical advantage for resource-constrained robotic platforms.

We rigorously evaluate GUIDE across diverse simulation environments with varying structural complexities and through real-world deployments on physical robotic platforms. Quantitative results demonstrate consistent improvements over state-of-the-art methods, with our approach achieving up to 18.3\% faster coverage completion and a 34.9\% reduction in redundant movements across benchmark environments. Qualitative analysis further showcases GUIDE's superior capability in structural inference and adaptive exploration behavior.
The main contributions of this work are threefold:

1) We introduce a region-evaluation global graph inference module that constructs a unified environmental representation by integrating observed information with predictions of unexplored areas. It incorporates a novel region-evaluation mechanism that assesses the reliability and decision relevance of predicted areas, enabling robust and reliable exploration planning under uncertainty.

2) We develop a diffusion-based decision-making framework that explicitly leverages the global graph representation to generate stable, long-horizon exploration trajectories, significantly reducing the computational burden of conventional diffusion policies while producing foresighted and efficient exploration paths.

3) We conduct comprehensive evaluations across multiple simulation environments and real-world scenarios, demonstrating GUIDE's superior performance in both structural inference precision and exploration efficiency, establishing a new benchmark for autonomous exploration systems.
\section{RELATED WORK}

Research on autonomous exploration is broadly divided into two paradigms. Model-based methods rely on explicitly designed planning schemes defined over environment representations to guide exploration, while learning-based methods use data-driven algorithms to learn exploration strategies from interaction or training. The following subsections review the developments in these two directions.
\subsection{Model-Based Exploration Methods}

\begin{figure*}[t] 
\vspace{5pt}
\centering
\includegraphics[width=0.8\textwidth]{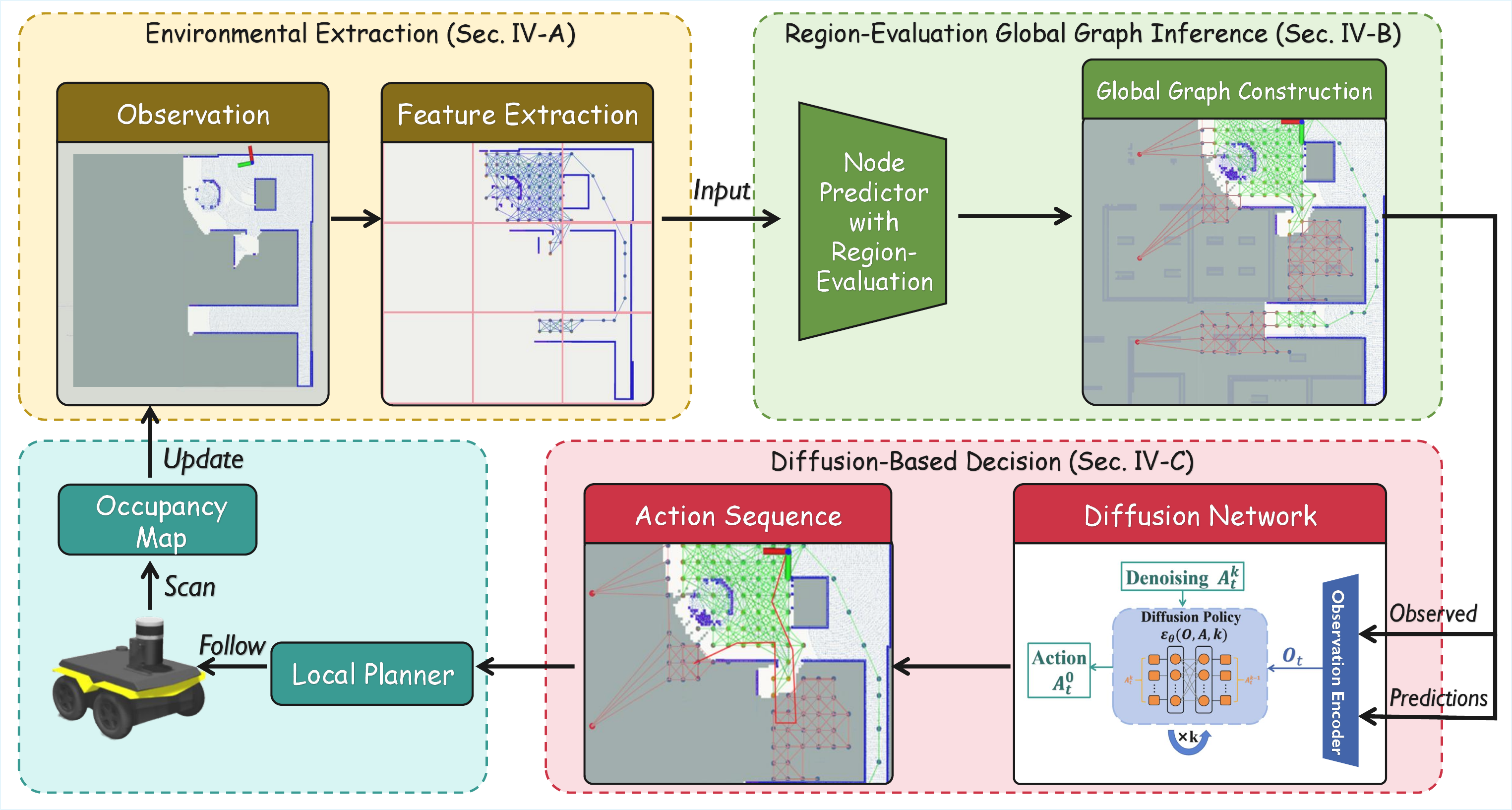} 
\caption{Overview of the proposed \textbf{GUIDE} framework with three modules:  
\textbf{(Sec.~IV-A)} Environmental Extraction updates the observation, samples free nodes, and decomposes the whole space;  
\textbf{(Sec.~IV-B)} Region-Evaluation Graph Inference leverages observations to predict nodes and filters them for reliability;  
\textbf{(Sec.~IV-C)} Diffusion-Based Decision conditions a diffusion policy on the global graph to generate stable long-horizon trajectories. }
\label{fig:pipeline}
\end{figure*}  

Classical exploration starts with frontier-based methods, first proposed by Yamauchi~\cite{yamauchi1997frontier}, which target the boundary between known and unknown areas. Sampling-based planners such as NBVP~\cite{nbvp} adopt a receding-horizon RRT scheme, but their local nature often results in myopic decisions and frequent revisits. Later extensions with history-aware strategies, motion primitives~\cite{mbplanner}, and graph-based formulations~\cite{dang2020graph} improve efficiency and consistency, yet still lack explicit long-horizon planning.

Hierarchical planning methods address this limitation by introducing a global visiting order. FUEL~\cite{zhou2021fuel} maintains an incremental frontier structure and solves an ATSP tour over frontier clusters to guide global exploration, reducing detours and improving efficiency. TARE~\cite{cao2021tare} coordinates sparse global routing with dense local refinement in a coarse-to-fine tree, preserving scalability and connectivity. Under sensing constraints, recent work improves viewpoint determination to curb revisits and wasted motion~\cite{zhao2022faep,yu2023echo}. However, these methods still rely exclusively on observed map information, making them inherently myopic toward unobserved areas and prone to inefficient planning and suboptimal coverage.

Coverage-path-based methods address this limitation by incorporating explicit coverage objectives to improve global path consistency and multi-robot coordination~\cite{feng2024fc,zhou2023racer}. Notably, FALCON~\cite{zhang2024falcon} exploits cues from unexplored areas to compute a global guidance path over unknown space, controlling local frontier visitation and suppressing backtracking. In the same spirit, dual-layer planners combine global region routing with local viewpoint and trajectory optimization to accelerate exploration while maintaining short and consistent tours~\cite{dong2025eden}. However, their reliance on uniform grid decomposition implicitly assumes environmental regularity, leading to performance degradation in spaces with irregular layouts or varying structural complexity.

\subsection{Learning-Based Exploration Methods}
Recent studies have explored learning-based methods to improve exploration under partially observed environments. Early work showed that reinforcement learning can acquire strategies directly from onboard sensing without explicit priors~\cite{tai2016robot,zhu2018deep}, though policies are often myopic. To mitigate this, graph-based methods introduced structural reasoning to manage uncertainty and support coordination~\cite{chen2020autonomous,hu2020voronoi}, while attention mechanisms capture multi-scale dependencies and predict information gain, enabling more consistent long-horizon planning in large-scale environments~\cite{cao2023ariadne,cao2024deep}. Subsequent advances include hierarchical decision networks~\cite{liang2024hdplanner}, dynamic graph methods~\cite{vashisth2024deep}, and hybrid schemes combining frontier heuristics~\cite{nam2024shangus}. More recently, diffusion-based policies generate complete trajectories for stable and coherent exploration~\cite{cao2024dare}, and multi-agent reinforcement learning with graph attention enables decentralized coordination under constrained sensing~\cite{chiun2025marvel}. While these methods improve adaptability, they remain constrained to encoding observed regions without modeling unexplored structures, fundamentally limiting long-horizon, globally efficient exploration.

Therefore, some works explicitly predict unobserved areas to provide guidance and mitigate myopic planning. Learned occupancy completion has been shown to improve information-gain estimation~\cite{shrestha2019learned,tao2022seer}, while over-reliance can lead to greedy and inefficient coverage~\cite{ericson2021understanding}. To mitigate this, recent works fine-tune LaMa~\cite{lama} on the KTH floor plan dataset~\cite{KTH} and use ensembles to model variance, yielding probabilistic information gain~\cite{ho2024mapex,baek2025pipe}. Although these methods enhance inference of unobserved structures, they are mostly applied to local planning and cannot fully exploit predictions for long-horizon optimization, while dense map representations further incur high computational cost.

\section{SYSTEM OVERVIEW}



We consider the task of robot exploration in a previously unknown environment. The robot incrementally constructs a 2D occupancy map $O_t$ from onboard sensor data, labeling each grid cell as free, occupied, or unknown. The planner is required to compute safe and feasible trajectories that expand $O_t$ over time until all reachable unexplored areas are revealed, thereby yielding a complete occupancy map of the accessible space while minimizing exploration cost, measured in distance traveled or time.


As illustrated in Fig.~\ref{fig:pipeline}, the proposed \textbf{GUIDE} consists of three modules. The \textbf{Environmental Extraction module} (Sec.~\ref{sec:environmental extraction}) updates the occupancy map $O_t$, samples free nodes, and decomposes the entire space into fixed-size regions. The \textbf{Region-Evaluation Global Graph Inference module} (Sec.~\ref{sec:graph_inference}) leverages observed information to predict nodes in unexplored areas, employing a region-evaluation mechanism to assess region reliability and filter predictions, thereby constructing a credible global graph representation. The \textbf{Diffusion-Based Decision module}  (Sec.~\ref{sec:diffusion}) conditions a diffusion policy directly on this global graph to generate foresighted action sequences with fewer denoising steps, producing stable long-horizon trajectories. The robot iteratively executes trajectories and updates $O_t$ with new sensor data, enabling exploration guided jointly by observations and region-evaluated inferences for improved efficiency and consistency.

\section{Methodology}


\subsection{Environmental Extraction}
\label{sec:environmental extraction}

Graph representations abstract the environment by modeling free positions as nodes and collision-free connections as edges. Unlike methods restricted to observed space, we construct a unified global graph that also includes placeholders for unexplored areas. Formally, the environment is represented as \(G=(V,E)\), where each node \(v \in V\) encodes its lattice coordinate at resolution \(d_n\), a type label (known/unknown), a visitation flag, neighborhood indices, and a utility score, which quantifies the node’s potential for exploration. From the occupancy map \(O_t\), we sample free cells to form the known free node set:
\begin{equation}
\mathcal{V}_f=\{\, (x,y)\mid O_t(x,y)=\text{free at resolution } d_n \,\},
\end{equation}
while the occupied cells form the obstacle node set \(\mathcal{V}_o\), which are used only as structural references.

To cover the entire environment, we decompose the space into fixed-size regions:
\begin{equation}
\mathcal{G}=\{ g_1,g_2,\dots,g_N \}.
\end{equation}
Each region is assigned one of the following labels: 

\textit{Explored}: if it contains observed free or occupied cells without frontiers.

\textit{Boundary}: if it contains frontier cells.

\textit{Unobserved}: if no cells are observed. For every unobserved region, its centroid is added to the grid node set \(\mathcal{V}_g\) as a placeholder.

\subsection{Region-Evaluation Global Graph Inference}
\label{sec:graph_inference}
\textbf{1) Global Node Predictor:}  
As outlined in Algorithm.~\ref{alg:node_pred}, the predictor estimates potential free nodes in unknown areas. First, the known free and obstacle nodes are rasterized onto a lattice with resolution $d_n$ to form a tri-valued image $I$ (255 free, 0 occupied, 128 unknown). The unknown pixels serve as a mask, which is completed by the LaMa inpainting model~\cite{lama}, fine-tuned on our node dataset, yielding $\hat I$ that predicts potential free nodes.

Candidate nodes are extracted from $\hat I$ by binarization to identify free nodes, then refined by removing isolated nodes through component analysis, yielding the final set $\mathcal{V}_p$.

\begin{algorithm}[htbp]
\caption{Global Node Predictor}
\label{alg:node_pred}
\KwIn{Known free nodes $\mathcal{V}_f$, obstacle nodes $\mathcal{V}_o$}
\KwOut{Predicted free nodes $\mathcal{V}_p$}

$I \leftarrow \textbf{Rasterize}(\mathcal{V}_f,\mathcal{V}_o,d_n)$\;
$\hat I \leftarrow \textbf{LaMa}(I,\text{mask}=128)$\;
$P \leftarrow \textbf{Binarize}(\hat I,\tau)$\;

$\mathcal{V}_p \leftarrow \textbf{Nodes}(P)$\;
\For{node $v \in \mathcal{V}_p$}{
    \If{\textbf{IsIsolated}($v$)}{remove $v$}
}

\Return $\mathcal{V}_p$\;
\end{algorithm}

\begin{figure}[htbp]
\vspace{5pt}
\centering
\includegraphics[width=0.48\textwidth]{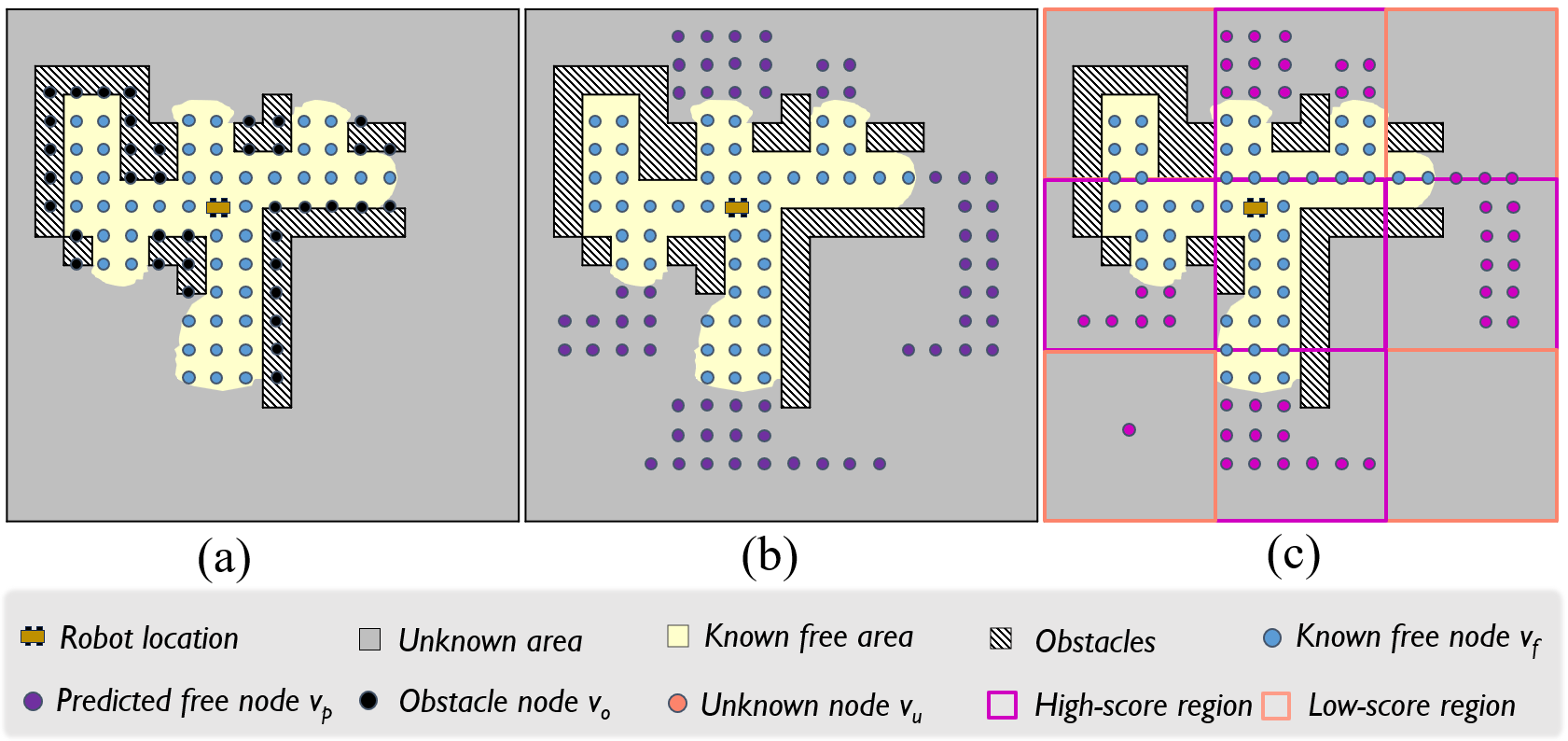} 
\caption{Illustration of the process of node prediction with region-evaluation.  
(a) Observed nodes (free and obstacles) as input to the predictor.  
(b) Predicted free nodes $\mathcal{V}_p$ from the Global Node Predictor.  
(c) Final unknown-node set $\mathcal{V}_u$ after region-evaluation, where high-score regions retain predictions while others are replaced by centroids or discarded.}
\label{fig:node}
\end{figure}

\textbf{2) Region-Evaluation Mechanism:}  
Based on the grid decomposition in Sec.~\ref{sec:environmental extraction}, we design a region-evaluation mechanism to assess regions and filter the predicted free nodes $\mathcal{V}_p$ for credible predictions (Fig.~\ref{fig:node}). Regions close to the robot and supported by frontiers are regarded as reliable for guiding exploration, while distant regions with limited information are less useful and represented sparsely.

Each region $g$ is assigned a score based on its centroid:
\begin{equation}
r(g) = \frac{\omega_f}{d_f} + \frac{\omega_r}{d_r},
\end{equation}
where $d_f$ is the average distance from the centroid to the $Z$ nearest frontiers and $d_r$ is the distance from the centroid to the robot. This scoring is computed only for regions labeled as \textit{Boundary} or \textit{Unobserved}.

We establish a region-evaluated unknown node set $\mathcal{V}_u$: for regions with higher scores, their predicted nodes from $\mathcal{V}_p$ are retained. 
In contrast, predictions in low-score regions are discarded. Specifically, in low-score \textit{Unobserved} regions, the region centroid $v_g$ is added to $\mathcal{V}_u$ when the predicted nodes within the region are sufficiently dense and discarded otherwise, to ensure a compact global representation.
The final node set is therefore
\begin{equation}
V = \mathcal{V}_f \cup \mathcal{V}_u,
\end{equation}
combining known free nodes with region-evaluated unknown nodes.

\begin{figure}[htbp]
\vspace{5pt}
\centering
\includegraphics[width=0.45\textwidth]{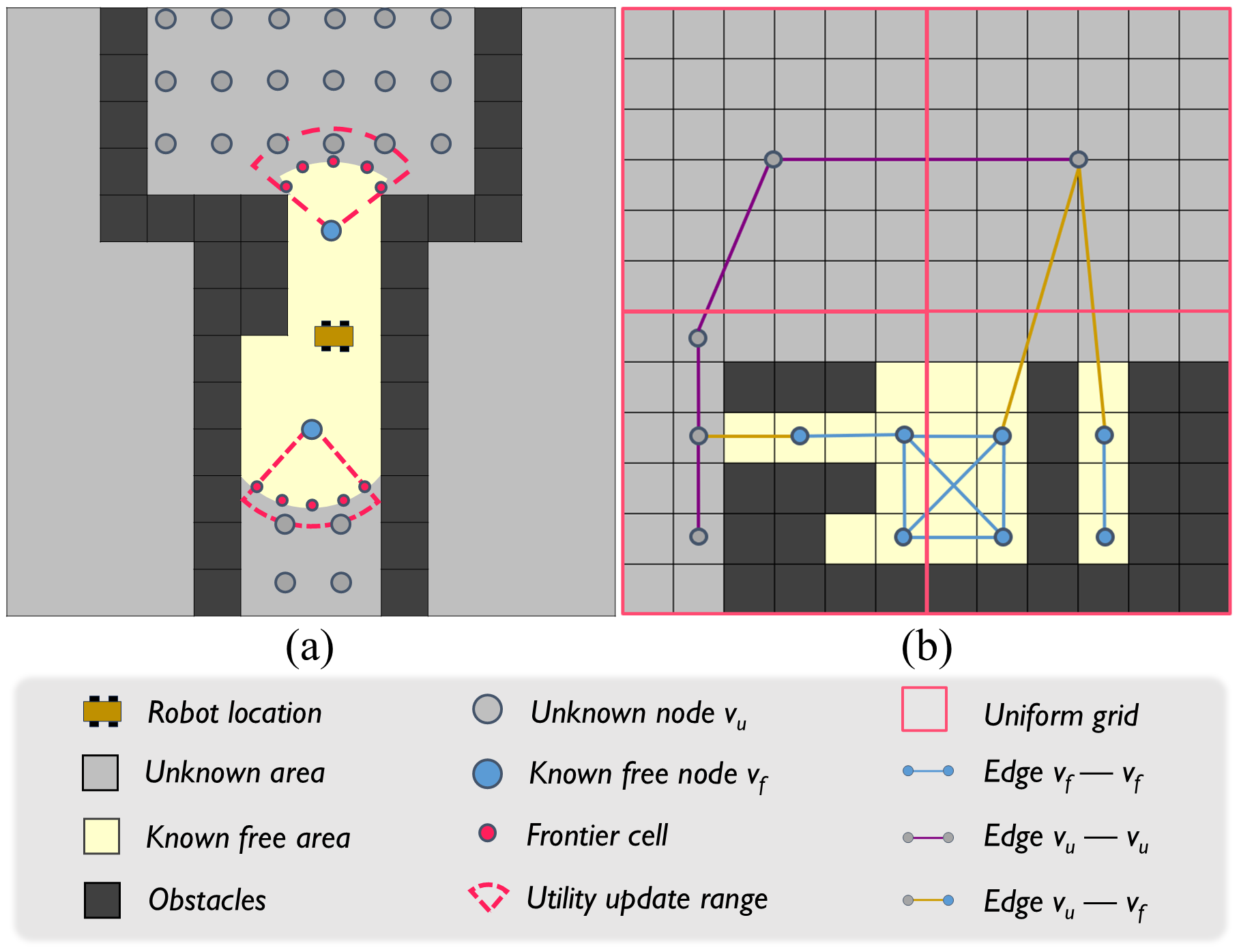} 
\caption{Illustration of (a) utility update, where nodes with the same frontier count obtain different utilities depending on predicted nodes within the utility update range; and (b) edge construction, showing connections between different node types.}
\label{fig:edge}
\end{figure}

\textbf{3) Global Graph Construction:}  
We update node utilities within a utility update range $r=\tfrac{R_{\text{sensor}}}{2}$ to measure exploration potential. As illustrated in Fig.~\ref{fig:edge}(a), frontier counts alone can be ambiguous—nodes with identical frontier numbers may correspond to vastly different unexplored extents. To resolve this, we incorporate predicted unknown nodes into the utility definition:
\begin{equation}
u(v)=
\begin{cases}
\min\!\Big\{F_{m}(v),\; F(v)\!\left(1+\dfrac{N_u(v)}{N_{um}(v)}\right)\Big\}, & v\in\mathcal{V}_f,\\[6pt]
0, & v\in\mathcal{V}_u,
\end{cases}
\end{equation}
where $F(v)$ is the number of frontier cells obtained by ray casting within the update range $r$, 
$N_u(v)$ is the number of predicted unknown nodes in this range, 
$N_{um}(v)$ is the maximum number of nodes that could occupy its unknown area at resolution $d_n$, 
and $F_m(v)$ is the upper bound on frontier count along the boundary of $r$. 
Thus, denser predicted nodes raise the utility while keeping it bounded by sensor visibility.

Edges are then constructed as shown in Fig.~\ref{fig:edge}(b). Free nodes connect to their collision-free neighbors, while unknown nodes are linked to nearby known or unknown nodes within the adjacent grid range, provided the path does not intersect occupied cells, even if it traverses unobserved space.

Combining observed free nodes with region-evaluated unknown nodes and their connecting edges, we obtain the global graph ${G} = ({\hat V}, {E} )$, which stores dynamically updated utilities and encodes feasible and hypothesized connectivity for exploration.

\subsection{Diffusion-Based Decision Guided by Global Graph}
\label{sec:diffusion}
We integrate the global graph from Sec.~\ref{sec:graph_inference} into the diffusion-policy framework. The graph $G$, which incorporates both observed information and predicted unexplored areas, is encoded by a Transformer encoder into a compact latent representation $z_t$. This graph embedding is then integrated into the observation to form an augmented observation $O_t$, which serves as the conditional input to the diffusion model.

At each timestep $t$, the action sequence $A^0_t = [a_t, a_{t+1} \dots, a_{t+T_p}]$ represents planned exploration moves within the prediction horizon $T_p$. Following the diffusion policy framework, an initial noisy action sequence $A^K_t$ is sampled from a Gaussian distribution and progressively denoised over $K$ iterations by a noise-prediction network $\epsilon_\theta$, yielding the final action sequence $A^0_t$. The denoising step is formulated as:
\begin{equation}
A^{k-1}_t = \alpha_k \left( A^k_t - \gamma_k \, \epsilon_\theta(O_t, z_t, A^k_t, k) \right) + \mathcal{N}(0, \sigma_k^2 I),
\end{equation}
where $\alpha_k, \gamma_k, \sigma_k$ are schedule-dependent parameters.  

Only the first $T_a$ steps are executed in a receding-horizon manner, ensuring reactivity to new information.  

The noise-prediction network $\epsilon_\theta$ is trained with expert demonstrations obtained from ground-truth trajectories. The training objective minimizes the mean squared error (MSE) between the sampled Gaussian noise $\epsilon_k$ and the corresponding model prediction:
\begin{equation}
\mathcal{L} = \mathbb{E}\left[ \| \epsilon_k - \epsilon_\theta(O_t, z_t, A^0_t + \epsilon_k, k) \|^2 \right].
\end{equation}

By leveraging the enriched global graph ${G}$ in the diffusion process, the policy generates stable and foresighted action sequence $A^0_t$ while substantially reducing the number of denoising steps $K$ compared to conventional diffusion policies.

Subsequently, the next waypoint is extracted from the generated action sequence $A^0_t$ and then directly provided to the local planner~\cite{cao2021tare}, which generates executable, collision-free trajectories.

\subsection{Training Details}
We adopt a two-stage (hierarchical) training scheme, where the first stage fine-tunes an inpainting network for node prediction and the second stage trains diffusion planner.

\textbf{Stage I:} The model is fine-tuned on 10,000 (observed, ground-truth) pairs collected from robot rollouts in randomly generated maze environments with sizes ranging from $100$\,m × $100$\,m to $180$\,m × $180$\,m. At multiple exploration steps, free and obstacle nodes are rasterized with a resolution of $d_n = 4$\,m, unknown masks are derived from observed information, and inputs are padded to sizes $32 \times 32$, $40 \times 40$, or $48 \times 48$.


\textbf{Stage II:} We keep the fine-tuned node prediction model fixed and train the diffusion planner conditioned on the global graph. 
We train the diffusion policy on 4,000 expert trajectories generated by a ground-truth-map coverage planner, where shortest coverage paths are computed via a Traveling Salesman Problem (TSP).
The diffusion network plans with horizon $T_p = 8$ given an observation horizon $T_o = 2$, executes within action horizon $T_a = 2$, and employs the Square Cosine Noise Scheduler~\cite{nichol2021improved} with $K = 30$ denoising steps.

Both stages are trained on a single RTX 4090, taking about 13 and 40 hours, respectively.



\section{EXPERIMENTS}

\subsection{Validation in Maze Datasets}
All experiments are conducted on a workstation equipped with an Intel i9-14900K CPU. We first evaluate the structural inference capability of our method, followed by comparisons with representative exploration baselines.

To evaluate the structural inference precision of our Global Graph inference method, we conduct tests in 100 unseen maze environments of size $100$\,m $\times$ $100$\,m. Precision is evaluated every 15 exploration steps as the proportion of region-evaluated 
unknown nodes $\mathcal{V}_u$ that fall within the ground-truth free space:
\begin{equation}
\mathrm{Pre}(k) = \frac{|\mathcal{V}_u^k \cap \mathcal{M}_{free}|}{|\mathcal{V}_u^k|}.
\end{equation}

As shown in Table~\ref{tab:baseline1}, our method can reliably predict potential free node locations. The high efficiency of the inference process is attributed to its lightweight representation, with each prediction completed in under $10$\,ms.

\begin{table}[htbp]
  \centering
  \caption{Average structural inference precision and prediction time over 100 mazes.}
  \label{tab:baseline1}
  \begin{tabular}{c|c|c|c|c}
    \toprule
    Steps ($k$) & 15 & 30 & 45 & 60 \\
    \midrule
    Precision (\%) & 75.92 & 79.73 & 79.24 & 84.61 \\
    Time (ms)     & 5.42  & 5.58  & 6.29  & 5.71  \\
    \bottomrule
  \end{tabular}
\end{table}

To further evaluate the exploration efficiency of our method, we compare it against several representative exploration planners, including NBVP~\cite{nbvp}, the DRL planner~\cite{cao2024deep}, MapEx~\cite{ho2024mapex}, the DARE planner~\cite{cao2024dare}, and the TARE local planner~\cite{cao2021tare}. For reference, we also provide an approximate optimal solution obtained from ground-truth coverage.

\begin{table}[htbp]
  \centering
  \caption{Comparison with baseline planners}
  \label{tab:baseline}
  \resizebox{\columnwidth}{!}{%
  \begin{tabular}{l|ccccc|c|c}
    \toprule
            & NBVP & DRL & MapEx & DARE & TARE  & Ours & Optimal \\
    \midrule
    Distance (\textit{m}) & 640(±105) & 575(±80) & 571(±82) & 565(±71) & 560(±65)  & 545(±65) & 501(±60) \\
    \midrule
    Gap to Optimal (\%) & 27.7\% & 14.8\% & 14.0\% & 12.8\% & 11.8\%  & \textbf{8.8\%} & 0\% \\
    \bottomrule
  \end{tabular}}
\end{table}

As shown in Table~\ref{tab:baseline}, we compare the average path lengths required by different methods to complete exploration. Compared with baseline methods, our approach achieves significantly shorter exploration paths. 

\begin{figure}[htbp]
\vspace{5pt}
\centering
\includegraphics[width=0.48\textwidth]{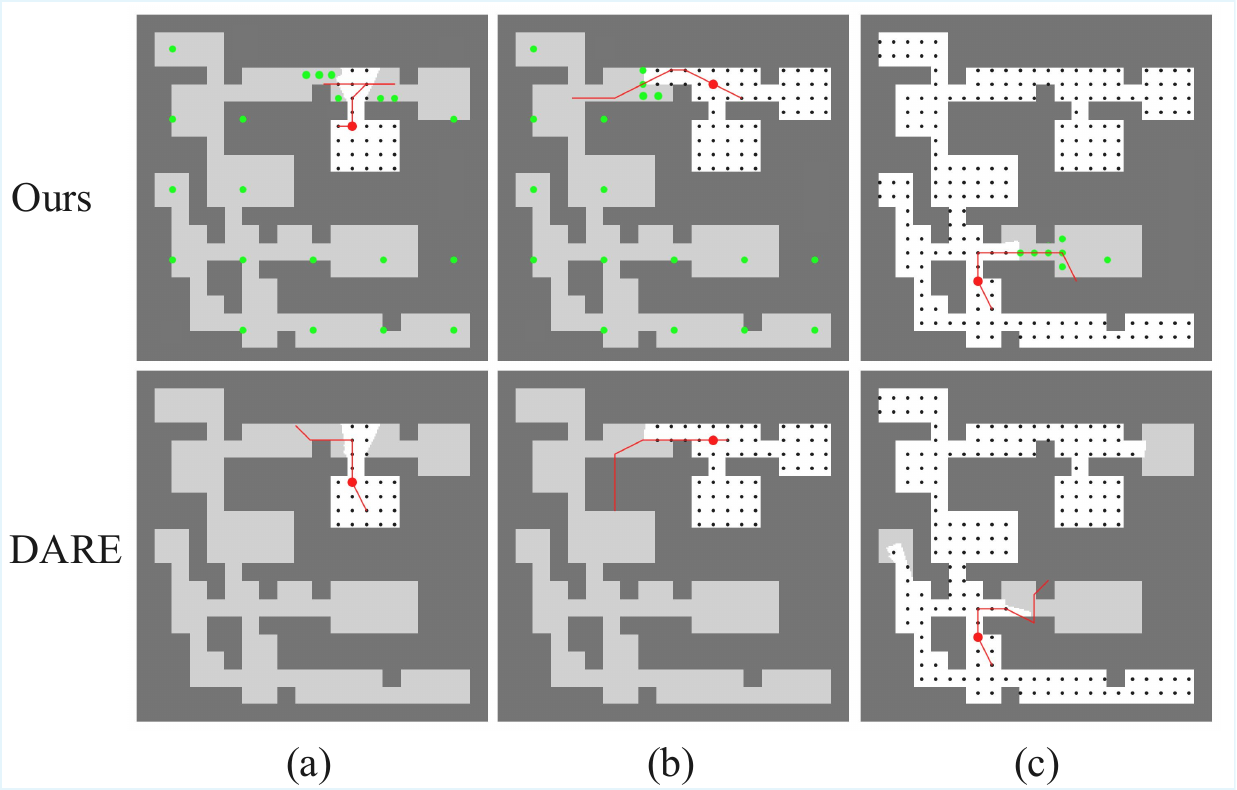} 
\caption{Action sequence on the same map at approximately the same locations (a)–(c). In each pair, the bottom shows diffusion based only on observed information and the top our method, with red indicating the generated action sequence and green the region-evaluated unknown nodes. By leveraging the global graph representation, our method produces more reasonable trajectories.}
\label{fig:diffusion}
\end{figure}

In particular, to highlight the contribution of the global graph representation to the diffusion model, we provide a comparison in Fig.~\ref{fig:diffusion}: unlike the diffusion model which relies on observed information, incorporating the global graph representation enables our approach to generate foresighted and efficient exploration paths across diverse scenarios.

\subsection{Ablation Experiments}
To examine the contribution of each major component in our framework, we conduct ablation studies focusing on the \textbf{Region-Evaluation Mechanism} and the \textbf{Diffusion-Based Decision framework}. To this end, we design the following variants for comparison:

(i) \textbf{Without Region-Evaluation Mechanism}: This variant makes decisions based on all predictions of unexplored areas without applying the region-evaluation mechanism.

(ii) \textbf{Without Diffusion}: This variant samples nodes from the global graph representation, and then generates exploration paths via TSP.

As shown in Table~\ref{tab:Ablation}, both variants result in noticeable performance drops compared to our full method, confirming that both the region-evaluation mechanism and diffusion-based decision-making framework are crucial for improving exploration efficiency.

\begin{table}[htbp]
  \centering
  \caption{Ablation Experiment}
  \label{tab:Ablation}
  \begin{tabular}{l|c|c}
    \toprule
    Method & Distance (\textit{m}) & Gap to Optimal (\%) \\
    \midrule
    Ours          & \textbf{545(±65)} & \textbf{8.8\%}  \\
    Ours(w/o Region-Evaluation)   & 561(±70)    & 11.9\%  \\         
    Ours(w/o Diffusion) & 580(±79)    & 15.7\% \\
    \midrule
    Optimal  & 501(±60)    & 0\% \\
    \bottomrule
  \end{tabular}
\end{table}

\subsection{Validation in Gazebo Environments}
To evaluate the exploration efficiency metric of our method in high-fidelity simulation, we conduct experiments in two structured Gazebo environments provided by~\cite{huang2023fael}, with map sizes of ($64$\,m × $85$\,m) and ($92$\,m × $74$\,m). The robot’s maximum velocity is $2$\,m/s, the map resolution to $0.4$\,m, the node resolution to $1.6$\,m, and the sensing range to $16$\,m.

\begin{table}[htbp]
  \centering
  \caption{RESULTS OF SIMULATIONS IN TWO INDOOR ENVIRONMENTS(5 RUNS EACH)}
  \label{tab:benchmark} 
  \begin{tabular}{l|c|c|c|c|c}
    \toprule
    
    scene & method & Distance & Time & Efficiency& Efficiency \\
    \&size(m)&&(\textit{m})&(\textit{s})&(\textit{m$^3$/m}))&(\textit{m$^3$/s})\\
    
    \midrule
                   & TARE & 1064.72 & 585.53 &5.11 &9.29 \\
              {Scene 1}       & DRL & 741.15 & 536.77 &7.34 &10.13  \\
              {64 × 85}       & DARE & 681.41 & 482.82 &7.98 &11.27  \\
                              & Ours & \textbf{537.18} & \textbf{432.11} &\textbf{10.13} &\textbf{12.59}  \\
    \midrule
                    & TARE & 1033.72 & 615.37 &6.58 &11.06  \\
             {Scene 2}        & DRL & 759.88 & 635.36 &8.96 & 10.72 \\
             {92 × 74}        & DARE & 797.38 & 623.27 &8.54 & 10.93 \\
                              & Ours & \textbf{531.37} & \textbf{513.09} &\textbf{12.81} &\textbf{13.27}  \\
    \bottomrule
  \end{tabular}
\end{table}

\begin{figure}[htbp]
\vspace{5pt}
\centering
\includegraphics[width=0.48\textwidth]{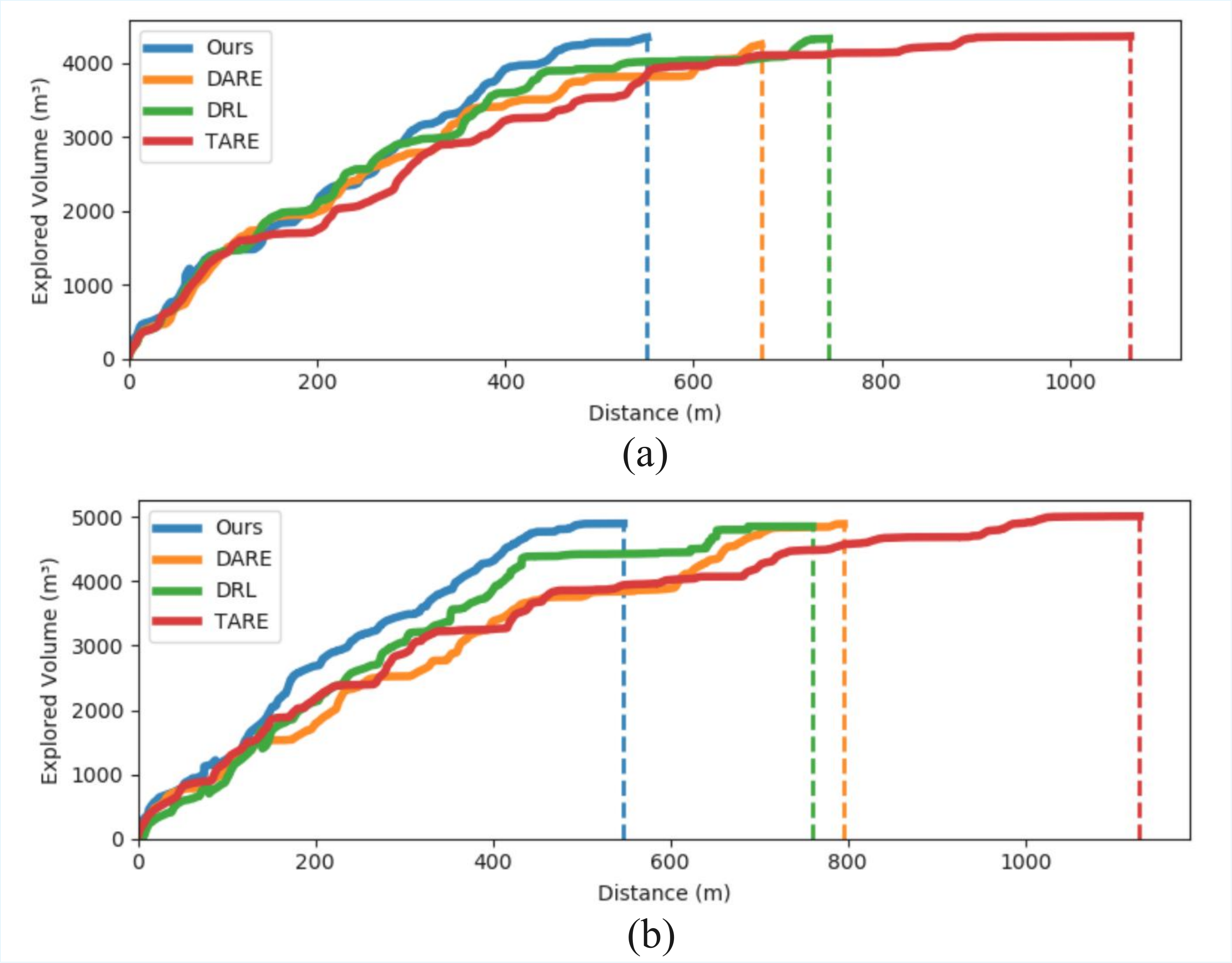} 
\caption{Comparison of exploration trajectories between our method and baseline planners in the Gazebo simulation: (a) Scene 1 and (b) Scene 2.}
\label{fig:figure}
\end{figure}

\begin{figure}[htbp]
\vspace{5pt}
\centering
\includegraphics[width=0.48\textwidth]{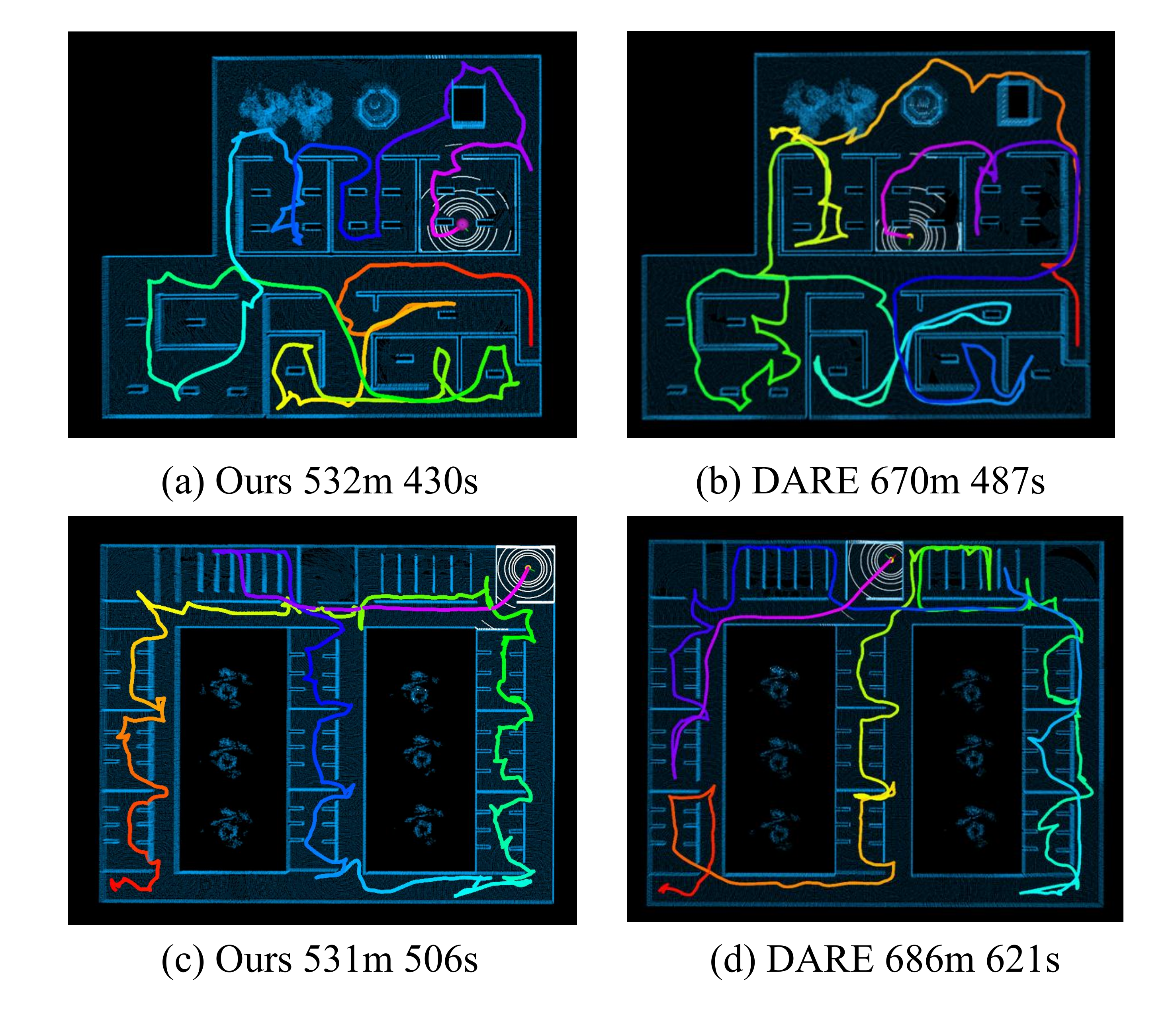} 
\caption{Comparison of exploration paths in the Gazebo simulation: (a)–(b) Scene 1 and (c)–(d) Scene 2.}
\label{fig:trajectory}
\end{figure}
 
Efficiency is defined as the explored free volume divided by the travel distance or the makespan. As shown in Table~\ref{tab:benchmark} and Fig.~\ref{fig:figure}, our method consistently achieves up to 18.3\% faster coverage and 34.9\% reduction in redundant movements compared with TARE local planner~\cite{cao2021tare}, DRL planner~\cite{cao2024deep}, and DARE planner~\cite{cao2024dare}. In addition, the comparative exploration paths illustrated in Fig.~\ref{fig:trajectory} demonstrate that our method generates foresighted and stable trajectories which maximize coverage efficiency while minimizing redundant movements. In environments with complex intersections and room layouts, the ability to predict the structure of unexplored areas effectively improves overall efficiency.

\begin{table}[htbp]
  \centering
  \caption{Comparison of average computation time and exploration efficiency under different denoising steps.}
  \label{tab:denoise_compare}
  \begin{tabular}{l|c|c|c|c}
    \toprule
    \multirow{2}{*}{Method} & \multicolumn{2}{c|}{$K=10$} & \multicolumn{2}{c}{$K=30$} \\
    \cmidrule(r){2-5}
     & Time (s) & Path (m) & Time (s) & Path (m) \\
    \midrule
    Ours   & \textbf{0.177} & \textbf{578.71} & \textbf{0.227} & \textbf{533.18} \\
    DARE   & 0.163 & 762.47 & 0.211 & 747.21 \\
    \midrule
    \multirow{2}{*}{Method} & \multicolumn{2}{c|}{$K=50$} & \multicolumn{2}{c}{$K=100$} \\
    \cmidrule(r){2-5}
     & Time (s) & Path (m) & Time (s) & Path (m) \\
    \midrule
    Ours   & \textbf{0.663} & \textbf{531.71} & \textbf{1.283} & \textbf{530.05} \\
    DARE   & 0.647 & 723.92 & 1.271 & 708.31 \\
    \bottomrule
  \end{tabular}
\end{table}

Furthermore, to demonstrate GUIDE’s ability to generate stable, foresighted action sequences with fewer denoising steps, we compare it with DARE~\cite{cao2024dare} under varying values of $K$. The experiments are conducted in the two structured Gazebo environments described above, and the results are reported as the averages across these two scenarios. As shown in Table~\ref{tab:denoise_compare}, DARE requires up to $K=100$ steps with the Scheduler~\cite{nichol2021improved} to produce stable trajectories, resulting in an average planning time of approximately $1.3$\,s. GUIDE leverages the enriched global graph representation to achieve reliable outputs with only $K=30$ denoising steps. This reduces the planning time to around $0.2$\,s while also yielding shorter overall exploration paths.

\subsection{Validation in Real World}

\begin{figure}[htbp]
\vspace{5pt}
\centering
\includegraphics[width=0.48\textwidth]{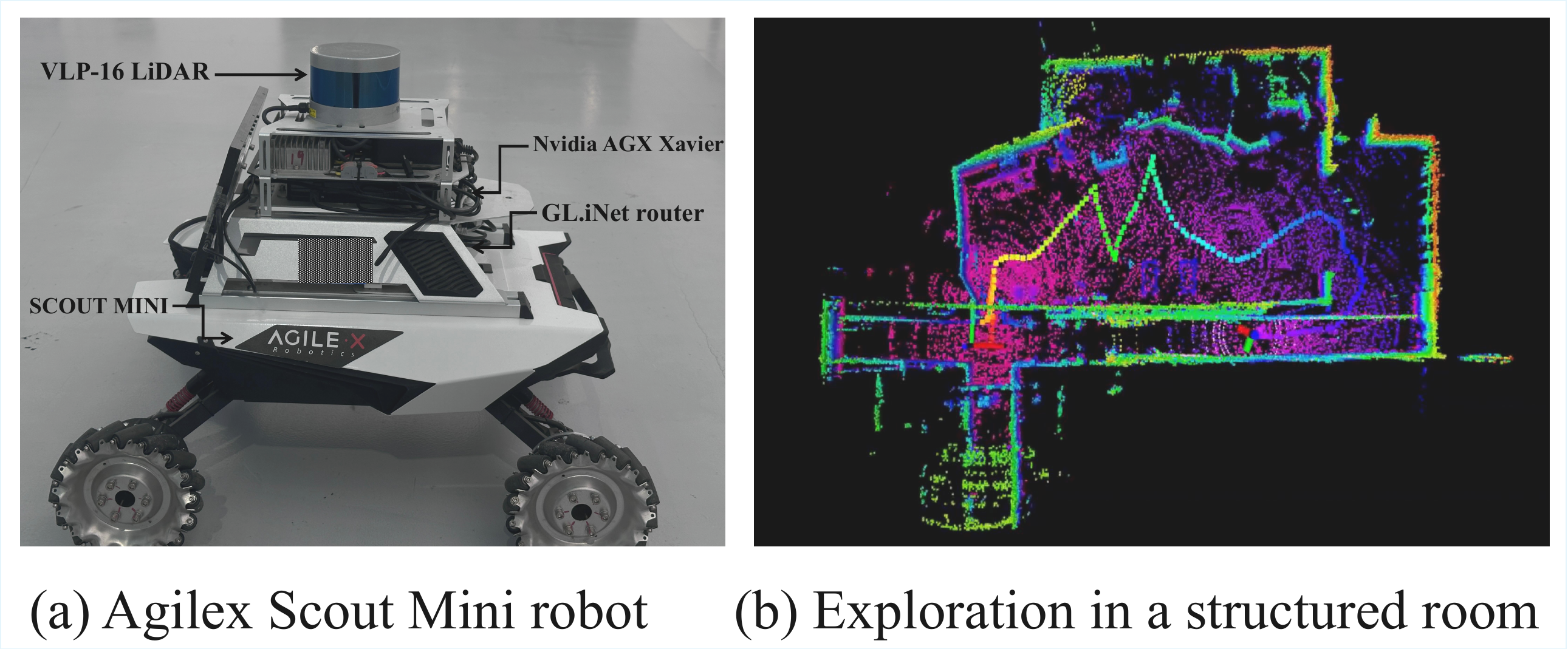} 
\caption{Real world experiment in a $40$\,m × $20$\,m structured environment.}
\label{fig:realexp}
\end{figure}

Finally, we validate our method in a structured real-world environment of size $40$\,m × $20$\,m, as shown in Fig.~\ref{fig:realexp}. The experiments are carried out using an Agilex Scout Mini robot equipped with a Velodyne VLP-16 LiDAR and an onboard NVIDIA AGX Xavier for computation. The robot operates with a maximum speed of $0.5$\,m/s, a map resolution of $0.2$\,m, a node resolution of $1$\,m, and a sensor range of $10$\,m during experiments. As shown in Fig.~\ref{fig:real}, the robot leverages the global graph representation to generate foresighted exploration trajectories. Results demonstrate that anticipating the structure of unknown areas effectively reduces redundant exploration and significantly improves overall efficiency, confirming the stability and practical applicability of GUIDE in real-world deployments.

\section{CONCLUSIONS}

In this work, we propose GUIDE, a novel autonomous exploration framework that integrates region-evaluation global graph inference with diffusion-based decision-making. By constructing a unified environmental representation that combines observed information with predictions of unexplored areas, GUIDE generates stable and foresighted exploration trajectories while requiring significantly fewer denoising steps. 
Extensive simulations and real-world deployments demonstrate GUIDE's superior performance in structural reasoning, coverage efficiency, and path optimality. Future work will focus on extending GUIDE to 3D environments and multi-robot systems to enable cooperative exploration and enhance scalability in larger and more complex scenarios.


\section*{ACKNOWLEDGMENT}

We would like to express our deepest gratitude to the Red Bird MPhil Program at The Hong Kong University of Science and Technology (Guangzhou) for providing the generous support, resources, and funding, which have been instrumental in the successful completion of our research.


\bibliographystyle{ieeetr}
\bibliography{ref}

\end{document}